\documentclass[runningheads]{llncs}
\usepackage[T1]{fontenc}
\usepackage{graphicx,verbatim}
\usepackage{amsmath}
\usepackage{hyperref}
\begin{document}
\title{Anatomical and Physical Supervision for CT-less PET Attenuation Correction: BIC-MAC 2026 Challenge}
\titlerunning{Anatomical \& Physical Supervision for CT-less PET Attenuation Correction}
%
\author{Petros Chatzitoulousis\inst{1} \and
George K. Matsopoulos\inst{1}}
\authorrunning{P. Chatzitoulousis and G.K. Matsopoulos}
\institute{$^{1}$Biomedical Engineering Laboratory, National Technical University of Athens, Athens, Greece}
  
\maketitle              
\begin{abstract}
This report describes our submission to the Big Cross-Modal
Attenuation Correction (BIC-MAC) 2026 Challenge for CT-less PET attenuation correction through multimodal pseudo-CT synthesis. We build upon a standard nnU-Net architecture and combine anatomical and physical supervision to improve both pseudo-CT quality and downstream PET reconstruction. Anatomical supervision is introduced through a frozen TotalSegmentator feature extractor, anatomy-guided structural constraints and patch sampling, while physical supervision is achieved using a differentiable attenuation correction factor projection loss based on multi-angle attenuation projections. Furthermore, the network is initialized with pretrained weights obtained from training on the SynthRAD Challenge MR-to-CT dataset. Minimal architectural modifications are applied, while performance improvements are pursued across the nnU-Net pipeline, including preprocessing, plans, and supervision design, among other components. Our final submission demonstrates the effectiveness of combining anatomical supervision, attenuation physics, and efficient nnU-Net scaling for CT-less PET attenuation correction.


\bigskip
\noindent\textbf{Correspondence:} pchatzi@biomed.ntua.gr

\end{abstract}
\section{Introduction}
Attenuation correction (AC) is a fundamental step in quantitative PET imaging and is conventionally performed using CT-derived attenuation maps \cite{chen2023,krokos2023}. In PET/MRI and dose-sensitive PET/CT applications, however, acquiring a volumetric CT is either unavailable or undesirable, motivating the synthesis of pseudo-CT images from alternative imaging modalities \cite{elkayee2026}. The BIC-MAC 2026 Challenge \cite{bicmac2026} addresses this problem by providing multimodal whole-body PET, MRI, and Topogram data and evaluates submissions primarily on the quantitative accuracy of reconstructed PET rather than pseudo-CT fidelity alone.

In this report, we present our submission to the BIC-MAC 2026 Challenge. Our approach builds upon the standard nnU-Net framework \cite{isensee2021} and its image synthesis extension, nnUNet\_translation \cite{longuefosse2024}. It further leverages the foundation established by the SynthRAD Challenge \cite{synthrad2023,synthrad2024}, where the MR-to-CT task provides the pretrained initialization for our model. We extend this framework to multimodal PET/MR-to-CT synthesis by incorporating NAC-PET and Topogram as additional input modalities, introducing attenuation physics supervision into the training objective, and optimizing the framework for the dual objectives of pseudo-CT synthesis and downstream PET attenuation correction. Rather than introducing major architectural changes, we explore performance improvements throughout the nnU-Net pipeline, including preprocessing, plans, and supervision design, while preserving the simplicity and reproducibility of the framework.

\section{Methods}
The proposed framework extends a standard nnU-Net for multimodal pseudo-CT synthesis, building upon its automatically configured preprocessing, network architecture, and experiment planning while introducing task-specific modifications throughout the training pipeline. The main components of the method are described in the following sections.

\subsection{Data and Preprocessing}
The proposed framework is trained on the multimodal BIC-MAC dataset, using NAC-PET, Dixon MRI (in- and out-of-phase), and the 2D Topogram as four input channels, with the co-registered CT volume as target. The Topogram is provided as a 2D scan-planning projection on the CT grid and is broadcast along the missing dimension to form a voxel-aligned fourth input channel, providing a 2D attenuation prior complementary to the volumetric modalities. Preprocessing follows the standard nnU-Net v2 pipeline, including automatic experiment planning, resampling to the target spacing, and per-channel intensity normalization. Each input channel is normalized independently by per-image z-scoring, while the CT target is normalized using fixed dataset-level statistics after Hounsfield Unit (HU) clipping. The network predicts normalized CT values that are converted back to HU during inference.

Provided anatomical annotations are used exclusively during training and are not required at inference. TotalSegmentator \cite{totalsegmentator2023} is used both as a frozen feature extractor for anatomical perceptual supervision and to provide organ labels for organ-aware attenuation projection supervision. Body and skull masks supervise body-wide gradient consistency and skull-specific structural consistency, respectively. Finally, anatomy-guided patch sampling increases the probability of selecting bone, soft tissue, skull, and organ regions, focusing training on anatomically relevant structures with high impact on PET attenuation correction.

\subsection{Network Architecture}

The proposed framework is built upon the standard PlainConvUNet architecture of nnU-Net v2, as implemented in the nnUNet\_translation repository. The network receives the four preprocessed input channels and predicts a single normalized CT volume. Apart from adapting the input/output channels and the loss function to the BIC-MAC task, the architecture remains largely unchanged, allowing the proposed improvements to originate from the training pipeline rather than network redesign.

\subsection{Loss Function}

\subsubsection{Anatomical Supervision}

Anatomical supervision combines voxel-wise synthesis with feature- and structure-based losses. A class-balanced weighted L1 loss assigns larger weights to attenuation-relevant regions, such as bone and soft tissue \cite{liu2021,spadea2021}, while reducing the influence of numerically dominant air/background voxels. Anatomical perceptual supervision is introduced through the Anatomy Feature Perception (AFP) loss \cite{longuefosse2025}, using a frozen TotalSegmentator network as a multi-scale feature extractor. An L1 loss between the predicted and reference CT feature maps encourages agreement in anatomical representations beyond voxel intensities. A body-wide gradient loss additionally matches spatial intensity gradients between the predicted and reference CT within the body region, encouraging the preservation of anatomical boundaries. Particular emphasis is placed on the skull due to its importance for brain attenuation correction. Within the skull mask, dedicated gradient-matching and intensity-dispersion terms are applied to preserve bone boundaries, structural details, and the distribution of predicted CT values.

\subsubsection{Physical Supervision}

Physical supervision is introduced through a differentiable attenuation correction factor (ACF) projection loss. Predicted and reference CT volumes are first converted from Hounsfield Units to linear attenuation coefficients ($\mu$) at 511 keV using the piecewise-linear transformation of Carney et al. \cite{carney2006}. Multi-angle line-integral projections are then computed from the resulting attenuation maps, approximating the attenuation information relevant to PET reconstruction without requiring computationally expensive iterative reconstruction during training. Projection consistency is enforced both globally, across the whole attenuation map, and locally using TotalSegmentator organ masks, where corresponding organ-wise projections are compared between prediction and target. The organ-aware term was evaluated using either all segmented structures or a focused subset of challenge-scored organs, with the focused-organ configuration selected for the final submission. This directly constrains attenuation properties relevant to quantitative PET.

\subsubsection{Total Loss}

The overall objective combines the voxel-wise synthesis, anatomical feature, body-gradient, and physical projection losses:
\begin{equation}
\mathcal{L}_{\mathrm{base}} =
0.7\mathcal{L}_{\mathrm{L1}}
+ 0.7\mathcal{L}_{\mathrm{AFP}}
+ 0.175\mathcal{L}_{\mathrm{bodygrad}}
+ 0.25\mathcal{L}_{\mathrm{ACF}} ,
\end{equation}
where $\mathcal{L}_{\mathrm{L1}}$ is the voxel-wise $L_1$ synthesis loss, $\mathcal{L}_{\mathrm{AFP}}$ is the anatomical feature perceptual loss, $\mathcal{L}_{\mathrm{bodygrad}}$ is the body-wide gradient consistency loss, and $\mathcal{L}_{\mathrm{ACF}}$ is the differentiable attenuation projection loss.

For patches containing the skull, the objective is augmented with the skull-specific gradient and intensity-dispersion terms:
\begin{equation}
\mathcal{L}_{\mathrm{skull}} =
\mathcal{L}_{\mathrm{base}}
+ 0.5\mathcal{L}_{\mathrm{skullgrad}}
+ 0.1\mathcal{L}_{\mathrm{skulldisp}} ,
\end{equation}
where $\mathcal{L}_{\mathrm{skullgrad}}$ is the skull-specific gradient consistency loss, and $\mathcal{L}_{\mathrm{skulldisp}}$ is the skull intensity-dispersion loss.

\subsection{Pretraining}

The network is initialized with weights pretrained on the SynthRAD MR-to-CT synthesis task using the same nnUNet\_translation backbone. The single SynthRAD MR volume is replicated across the four input channels to match the BIC-MAC architecture, allowing all pretrained weights, including the input layer, to be transferred directly. Pretraining follows the pipeline described above, using the class-balanced weighted L1, frozen-TotalSegmentator AFP, and body-wide gradient losses. As SynthRAD contains neither PET nor Topogram data, ACF-projection supervision and Topogram conditioning are disabled during this stage. The resulting checkpoint initializes training on the multimodal BIC-MAC data, where the complete anatomical and physical supervision and all four input modalities are activated for fine-tuning.

\subsection{Training Configuration and Scaling}

Training follows the standard nnU-Net configuration, using SGD with Nesterov momentum, polynomial learning-rate decay, mixed-precision training, and 1000 epochs. Performance scaling is explored through nnU-Net plans by varying patch and batch sizes while preserving the underlying network topology and compatibility with the pretrained weights. Larger patches increase the spatial context available to the network and provide longer projection paths for the ACF loss. The final configuration uses a $192^3$ patch size with a batch size of 2. Larger batch sizes were not combined with $192^3$ patches due to the substantial memory requirements of the multi-scale anatomical and multi-angle projection supervision.

\section{Preliminary Results}
The proposed model was evaluated on the official BIC-MAC validation set using the four challenge metrics: CT attenuation-map MAE ($\mu$ at 511~keV), whole-body SUV MAE, organ bias, and brain outlier score. All metrics are reported such that lower values indicate better agreement with the respective reference. Results correspond to a single model (single fold), applied without test-time augmentation, and are summarized in Table~\ref{tab:results}.

\begin{table}
\caption{Performance on the BIC-MAC validation set across the four challenge metrics. For all metrics, lower is better ($\downarrow$).}\label{tab:results}
\centering
\setlength{\tabcolsep}{8pt}
\begin{tabular}{|l|l|r|}
\hline
\textbf{Domain} & \textbf{Metric} & \textbf{Value} \\
\hline
CT  & $\mu$-map MAE $\downarrow$ & 0.00572 \\
PET & Whole-body SUV MAE $\downarrow$ & 0.0367 \\
PET & Organ Bias (\%) $\downarrow$ & 2.70 \\
PET & Brain Outlier Score $\downarrow$ & 0.0136 \\
\hline
\end{tabular}
\end{table}

The model produces pseudo-CT volumes with attenuation coefficients closely matching the reference, while maintaining low errors in downstream PET quantification. Whole-body SUV MAE indicates good agreement in global tracer uptake, while the low organ bias suggests consistent recovery of mean uptake across the evaluated organs without pronounced over- or under-estimation in individual structures. The brain outlier score is particularly low and ranks among the best results on the official validation set, indicating reliable PET quantification in a region especially sensitive to errors in skull attenuation. Overall, the results demonstrate consistent performance across both CT synthesis and downstream PET evaluation.

\section{Discussion}

The results suggest that pseudo-CT synthesis for PET attenuation correction benefits from considering both anatomical accuracy and the physical role of the predicted CT in PET reconstruction. The combination of voxel-wise, feature-based, and structural supervision with attenuation projection constraints produced consistent performance across both CT- and PET-based metrics. Particularly strong performance was observed for the brain outlier metric, which was also among the metrics most improved by transfer learning from the SynthRAD MR-to-CT task during model development. This suggests that pretraining on a larger MR-to-CT synthesis dataset provides useful anatomical initialization, particularly for the representation of the skull and surrounding structures that are important for brain attenuation correction. At the same time, the approach retains the standard nnU-Net architecture and concentrates task-specific modifications on supervision, pretraining, and configuration scaling, providing a relatively simple and reproducible framework for multimodal attenuation correction. Further evaluation on the hidden BIC-MAC test set will be required to assess the generalization of these findings.

%
%
%

\begin{thebibliography}{8}

\bibitem{chen2023}
Chen, X., Liu, C.: Deep-learning-based methods of attenuation correction for SPECT and PET. Journal of Nuclear Cardiology \textbf{30}(5), 1859--1878 (2023). \doi{10.1007/s12350-022-03007-3}

\bibitem{krokos2023}
Krokos, G., MacKewn, J., Dunn, J., Marsden, P.: A review of PET attenuation correction methods for PET-MR. EJNMMI Physics \textbf{10}(1), 52 (2023). \doi{10.1186/s40658-023-00569-0}

\bibitem{elkayee2026}
Elkayee Dehno, A., Ghafarian, P., Arabi, H., Ay, M.R.: Simultaneous attenuation and scatter correction of PET data in the image: quantitative and clinical assessment of image-to-image deep learning models. Physica Medica \textbf{141}, 105683 (2026). \doi{10.1016/j.ejmp.2025.105683}

\bibitem{bicmac2026}
Hinge, C., Ladefoged, C.N., Andersen, F.L., Schramm, G., Korsholm, K., Law, I.: BIC-MAC: Big Cross-Modal Attenuation Correction Challenge. MICCAI 2026 Challenge (2026). \doi{10.5281/zenodo.19731820}

\bibitem{isensee2021}
Isensee, F., Jaeger, P.F., Kohl, S.A.A., Petersen, J., Maier-Hein, K.H.: nnU-Net: a self-configuring method for deep learning-based biomedical image segmentation. Nature Methods \textbf{18}(2), 203--211 (2021).
\doi{10.1038/s41592-020-01008-z}

\bibitem{longuefosse2024}
Longuefosse, A., Le Bot, E., Denis de Senneville, B., Giraud, R.,
Mansencal, B., Coupé, P., Desbarats, P., Baldacci, F.:
Adapted nnU-Net: A robust baseline for cross-modality synthesis and
medical image inpainting. In: Simulation and Synthesis in Medical Imaging,
LNCS, vol. 15187, pp. 24--33. Springer, Cham (2024).
\doi{10.1007/978-3-031-73281-2_3}

\bibitem{synthrad2023}
Thummerer, A., van der Bijl, E., Galapon Jr., A., Verhoeff, J.J.C.,
Langendijk, J.A., Both, S., van den Berg, C.N.A.T., Maspero, M.:
SynthRAD2023 Grand Challenge dataset: Generating synthetic CT for radiotherapy.
Medical Physics \textbf{50}(7), 4664--4674 (2023).
\doi{10.1002/mp.16529}

\bibitem{synthrad2024}
Huijben, E.M.C., Terpstra, M.L., Galapon Jr., A., Pai, S., Thummerer, A.,
Koopmans, P., Afonso, M., et al.: Generating synthetic computed tomography
for radiotherapy: SynthRAD2023 challenge report.
Medical Image Analysis \textbf{97}, 103276 (2024).
\doi{10.1016/j.media.2024.103276}

\bibitem{totalsegmentator2023}
Wasserthal, J., Breit, H.C., Meyer, M.T., Pradella, M., Hinck, D.,
Sauter, A.W., Heye, T., Boll, D.T., Cyriac, J., Yang, S., Bach, M.,
Segeroth, M.: TotalSegmentator: Robust segmentation of 104 anatomic
structures in CT images. Radiology: Artificial Intelligence
\textbf{5}(5), e230024 (2023). \doi{10.1148/ryai.230024}

\bibitem{liu2021}
Liu, Y., Lei, Y., Wang, Y., Shafai-Erfani, G., Wang, T., Tian, S.,
Patel, P., Jani, A.B., McDonald, M., Curran, W.J., Liu, T., Yang, X.:
Abdominal synthetic CT generation from MR Dixon images using a U-net
trained with ``semi-synthetic'' CT data. Physics in Medicine \& Biology
\textbf{66}(12), 125001 (2021).
\doi{10.1088/1361-6560/abfef2}

\bibitem{spadea2021}
Spadea, M.F., Maspero, M., Zaffino, P., Seco, J.:
Deep learning based synthetic-CT generation in radiotherapy and PET:
A review. Medical Physics \textbf{48}(11), 6537--6566 (2021).
\doi{10.1002/mp.15150}

\bibitem{longuefosse2025}
Longuefosse, A., Denis de Senneville, B., Dournes, G., Benlala, I.,
Baldacci, F., Desbarats, P.:
Anatomical feature-prioritized loss for enhanced MR to CT translation.
Physics in Medicine \& Biology \textbf{70}(14), 145012 (2025).
\doi{10.1088/1361-6560/adea07}

\bibitem{carney2006}
Carney, J.P.J., Townsend, D.W., Rappoport, V., Bendriem, B.:
Method for transforming CT images for attenuation correction in PET/CT imaging.
Medical Physics \textbf{33}(4), 976--983 (2006).
\doi{10.1118/1.2174132}

\end{thebibliography}
%

\end{document}